\documentclass[sigconf]{acmart}

\usepackage{multirow}
\usepackage{subfig}
\usepackage{enumerate}

\usepackage{algorithm}
\usepackage{algorithmic}

\usepackage{amsmath}

\usepackage{pifont}
\newcommand{\cmark}{\ding{51}}%
\newcommand{\xmark}{\ding{55}}%

\usepackage{ragged2e}

\usepackage{stfloats}
\AtBeginDocument{%
  \providecommand\BibTeX{{%
    \normalfont B\kern-0.5em{\scshape i\kern-0.25em b}\kern-0.8em\TeX}}}

\setcopyright{acmlicensed}

\copyrightyear{2025}
\acmYear{2025}
\setcopyright{acmlicensed}\acmConference[KDD '25]{Proceedings of the 31st ACM
SIGKDD Conference on Knowledge Discovery and Data Mining V.2}{August 3--7,
2025}{Toronto, ON, Canada}
\acmBooktitle{Proceedings of the 31st ACM SIGKDD Conference on Knowledge Discovery and Data Mining V.2 (KDD '25), August 3--7, 2025, Toronto, ON, Canada}
\acmDOI{10.1145/3711896.3737004}
\acmISBN{979-8-4007-1454-2/2025/08}
\begin{document}

\title{Improving Open-world Continual Learning under the Constraints of Scarce Labeled Data}

\author{Yujie Li}
\email{liyj1201@gmail.com}
\affiliation{%
  \institution{Southwestern University of Finance and Economics}
  \city{Chengdu}
  \country{China}
  \\
  \institution{Leiden University}
  \city{Leiden}
  \country{Netherlands}
}

\author{Xiangkun Wang}
\email{xiangkunwang18@gmail.com}
\affiliation{%
  \institution{Southwestern University of Finance and Economics}
  \city{Chengdu}
  \country{China}
}

\author{Xin Yang}
\authornote{Corresponding author.}
\email{yangxin@swufe.edu.cn}
\affiliation{%
  \institution{Southwestern University of Finance and Economics}
  \city{Chengdu}
  \country{China}
}

\author{Marcello Bonsangue}
\email{m.m.bonsangue@liacs.leidenuniv.nl}
\affiliation{%
  \institution{Leiden University}
  \city{Leiden}
  \country{Netherlands}
}

\author{Junbo Zhang}
\email{msjunbozhang@outlook.com}
\affiliation{%
  \institution{JD Intelligent Cities Research}
  \city{Beijing}
  \country{China}
}

\author{Tianrui Li}
\email{trli@swjtu.edu.cn}
\affiliation{%
  \institution{Southwest Jiaotong University}
  \city{Chengdu}
  \country{China}
}

\renewcommand{\shortauthors}{Yujie Li et al.}
\begin{abstract}
Open-world continual learning (OWCL) adapts to sequential tasks with open samples, learning knowledge incrementally while preventing forgetting. 
However, existing OWCL still requires a large amount of labeled data for training, which is often impractical in real-world applications. 
Given that new categories/entities typically come with limited annotations and are in small quantities, a more realistic situation is OWCL with scarce labeled data, i.e., few-shot training samples.
Hence, this paper investigates the problem of open-world few-shot continual learning (OFCL), challenging in (i) learning unbounded tasks without forgetting previous knowledge and avoiding overfitting, (ii) constructing compact decision boundaries for open detection with limited labeled data, and (iii) transferring knowledge about knowns and unknowns and even update the unknowns to knowns once the labels of open samples are learned.
In response, we propose a novel OFCL framework that integrates three key components: (1) an instance-wise token augmentation (ITA) that represents and enriches sample representations with additional knowledge, (2) a margin-based open boundary (MOB) that supports open detection with new tasks emerge over time, and (3) an adaptive knowledge space (AKS) that endows unknowns with knowledge for the updating from unknowns to knowns. Finally, extensive experiments show that the proposed OFCL framework outperforms all baselines remarkably with practical importance and reproducibility. The source code is released at \url{https://github.com/liyj1201/OFCL}.
\end {abstract}




\begin{teaserfigure}\centering
  \includegraphics[width=0.775\textwidth]{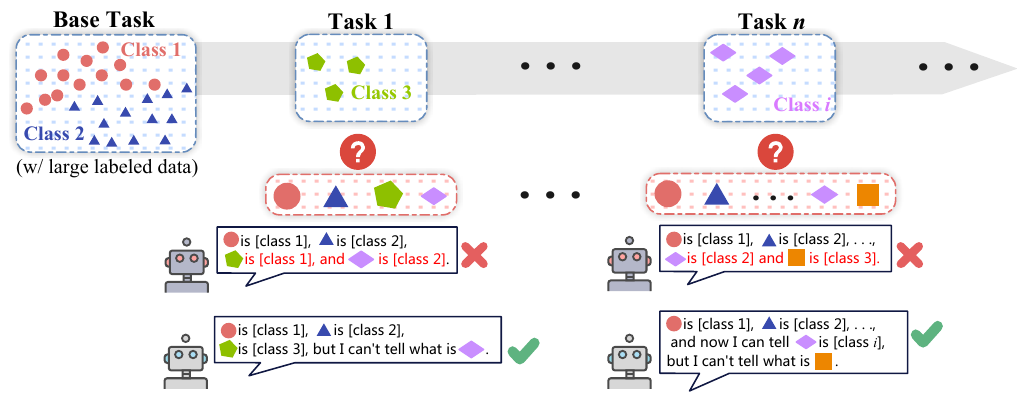}
  \vspace{-3mm}
  \caption{\textbf{The illustration of open-world continual learning with few-shot data (OFCL) framework.} As new tasks (Task 1, ..., Task $n$, etc.) are introduced with new classes (Class 3, ..., Class $i$, etc.) in few-shot labeled examples, unknown samples may appear during testing and the learner struggles with severer forgetting and overfitting. Hence, the OFCL requires the agent to detect open samples in testing and learn new categories from scarce labeled data over time.}
  \label{fig:intro}
\end{teaserfigure}

\maketitle

\section{Introduction}\label{sec:intro}
Recently, open-world continual learning (OWCL)~\cite{kim2022theoretical,liu2023ai} is an emerging machine learning paradigm where a model continuously learns from a sequence of tasks/data in an open environment where unseen classes may emerge during tests over time. 
Unlike traditional continual learning (CL) that operates in a closed and fixed set of classes (i.e., the closed-world assumption), OWCL requires the model to not only learn new knowledge and update knowledge incrementally but also to recognize and adapt to open, previously unseen classes without forgetting~\cite{bendale2015towards}.
Therefore, OWCL provides a more comprehensive formulation of real-world scenarios by involving the presence of open classes during tests, making it more suitable for managing the complexities of open data content~\cite{liu2023ai,li2024learning}.

Despite the increasing attention towards OWCL research, existing studies~\cite{joseph2021towards,kim2023open} face a major limitation due to their reliance on large amounts of training data for learning each new task. This requirement for abundant training samples, however, often contradicts the nature of the open and dynamic data, where practical applications usually encounter new tasks (or classes) with scarce labeled data or few-shot training samples~\cite{ahmad2022few,yoon2023soft}.
For instance, on social media platforms, user-uploaded images (e.g., user-generated photos or social media posts) often feature novel objects or scenes that lack corresponding labeled data. Unfortunately, obtaining sufficient manual and ground-truth labels for training each new task or class is often impractical. Hence, current OWCL models must be extended to operate with scarce labeled examples.

Therefore, motivated by the demands of real-world applications, we investigate the problem of OWCL under the restrictions of scarce labeled data, i.e., \textbf{O}pen-world \textbf{F}ew-shot \textbf{C}ontinual \textbf{L}earning (OFCL).
As depicted in \autoref{fig:intro}, the scarce labeled data exacerbates the issues of misclassifying open samples into seen classes and catastrophic forgetting (CF) due to insufficient knowledge representation.
Accordingly, OFCL faces the following two key challenges:
(1) \textit{Open Detection}: The few-shot training/labeled data in OWCL increases the difficulty of establishing robust boundaries for open detection.
(2) \textit{Knowledge Transfer}: Given the scarce labeled data, the learner not only needs to transfer knowledge by incrementally acquiring new knowledge without CF but also be able to update the knowledge of open samples to known classes once the labels of open samples are learned.
Therefore, to address these challenges, we propose a novel framework, termed \textbf{OFCL}, designed to tackle open-world continual learning with scarce labeled data. 

Specifically, we first propose an instance-wise token augmentation (ITA) aimed at acquiring additional `knowledge' to mitigate the inadequate representation caused by scarce labeled data. 
Moreover, ITA can facilitate knowledge transfer by matching learnable tokens to each sample embedding.
Additionally, given the scarce labeled data where certain exemplar points (hubs) appear among the nearest neighbors of many other points, test samples will be assigned to it regardless of their true label, resulting in low accuracy \cite{fei2021z}. 
To mitigate this, inspired by the embedding representations on a hypersphere \cite{trosten2023hubs}, we introduce a novel and compact margin-based open boundary (MOB) and an adaptive knowledge space (AKS) consisting of learnable hyperspheres, where each hypersphere is characterized by a class centroid and an associated radius. 
In particular, enables the formulation of compact decision boundaries between known and unknown samples, thereby enhancing open detection.
Simultaneously, the AKS encourages the model to learn incrementally from unknowns and classify previously encountered unknown samples in new tasks, effectively transforming unknowns into knowns over time. Extensive experimental results demonstrate that the proposed novel OFCL framework significantly surpasses all baseline methods, highlighting its practical significance and reproducibility.

In summary, our contributions can be outlined as follows:
\begin{itemize}
    \item Driven by practical applications, this paper formulates the problem of open-world continual learning with scarce labeled data and tries to address the challenges of open detection and knowledge transfer with limited ground-truth data.
    \item Technically, we introduce instance-wise token augmentations (ITA) for enhancing the semantic information of embeddings and mitigating forgetting. Subsequently, we propose a margin-based open boundary (MOB) and an adaptive knowledge space (AKS) to incorporate knowledge learned from knowns and unknowns, fostering knowledge transfer, accumulation, and even updating the unknowns.
    \item Extensive comparisons with competitive OWCL approaches, few-shot incremental learning methods and open detection baselines demonstrate the superior performance of our OFCL framework. Additional studies also show the effectiveness and robustness of the proposed OFCL framework.
\end{itemize}

\section{Related Work}

\subsection{Open-world Continual Learning (OWCL)}
In contrast to the closed-world assumption, open-world continual learning aims to reject examples from unseen classes (not appearing in training) and incrementally learn the open/unseen classes \cite{bendale2015towards,joseph2021towards}, similar to learning in the real world that is full of unknowns or novel objects \cite{liu2023ai,10.1145/3581783.3612011}.
The innovative framework SOLA \cite{liu2023ai} enables AI agents to learn and adapt by themselves via their interactions with humans and the open environment. 
PointCLIP \cite{zhu2023pointclip} combines CLIP and GPT-3 to enable zero-shot classification, segmentation, and open-set detection. 
Zhao et al.~\cite{zhao2023revisiting} introduce an Open World Object Detection (OWOD) framework containing an auxiliary proposal advisor and a Class-specific Expelling Classifier (CEC). \cite{kim2023open} proposes to integrate out-of-distribution detection with continual learning for open-world continual learning.

However, as mentioned in \autoref{sec:intro}, more and more real-world scenarios pose a greater challenge due to the limitation of extremely scarce training/labeled samples. Consequently, addressing open-world learning in scarce labeled data becomes not only highly practical but also increasingly challenging.

\begin{table}[ht]
\centering
\begin{tabular}{ccccc}
\toprule
\textit{\textbf{CL Baselines}} & \textbf{Open-world Assumption} & \textbf{Few-shot Data} \\ \midrule
\begin{tabular}[c]{@{}c@{}}TOPIC \cite{tao2020few} \end{tabular} & \xmark  & \cmark \\ \midrule
ALICE \cite{peng2022few}& \xmark & \cmark \\ 
\midrule
SoftNet \cite{yoon2023soft} & \xmark & \cmark \\ 
\midrule
SOLA \cite{liu2023ai} & \cmark & \xmark\\ 
\midrule
CEC~\cite{zhao2023revisiting} & \cmark & \xmark\\ 
\midrule
Pro-KT \cite{li2024learning} & \cmark & \xmark\\ 
\midrule
FeSSSS \cite{ahmad2022variable} & \cmark & \cmark\\  
\bottomrule
\end{tabular}
\caption{Competitive CL Baselines (selected) under Open-World Assumption and Few-Shot Data Constraints.}\label{tab:survey}
\vspace{-0.5cm}
\end{table}

\subsection{Few-shot Learning}
Few-shot learning~\cite{tao2020few,zhou2022forward,peng2022few} aims to learn new tasks with limited training data while avoiding the loss of previously learned information. 
TOPIC \cite{tao2020few} introduces the few-shot learning with continual learning benchmark setup and proposes a neural gas structure to balance between the past and the current tasks. 
SvF \cite{zhao2021mgsvf} employs frequency-aware regularization and feature space composition for a balanced approach. 
CEC \cite{zhang2021few} decoupled feature extraction and classification through a pre-trained backbone and a non-parametric class mean classifier. 
SoftNet \cite{yoon2023soft} utilizes a partial-frozen strategy, selectively masking a sub-network from a pre-trained network to prevent forgetting.

While current methods yield satisfactory results, they remain rooted in the closed-world assumption, lacking the capability to identify and address open/unknown samples during test phases \cite{10.1145/3581783.3612071,li2024ruleprompt}. Furthermore, there is a lack of methods for constructing a comprehensive knowledge space that facilitates knowledge accumulation, transfer, and updating throughout the incremental learning process.
Hence, given the open-world assumption, existing methods suffer even more from overfitting and CF, often leading to misinterpretation or revision of incorrect knowledge.

More recently (as shown in \autoref{tab:survey}), \cite{ahmad2022variable,10354388} tried to rethink the existing few-shot incremental learning under the open-set hypothesis. 
FeSSSS \cite{ahmad2022variable,ahmad2022few} extended a variable-few-shot learning model to the open world, detecting unknown samples by a simple Softmax thresholding approach (MSP \cite{hendrycks2016baseline}).
Hyper-RPL \cite{10354388} utilized the hyperbolic reciprocal point learning into a distillation-based framework to alleviate the overfitting and forgetting.
However, although both works made attempts, they merely combined existing few-shot incremental learning with simple open-set recognition methods, overlooking the challenges posed by scarce labeled data in OWCL and facing significant limitations in constructing compact open-set boundaries and adapting to dynamic environments.

\begin{figure*}[ht]
\centering
\includegraphics[width=0.95\textwidth]{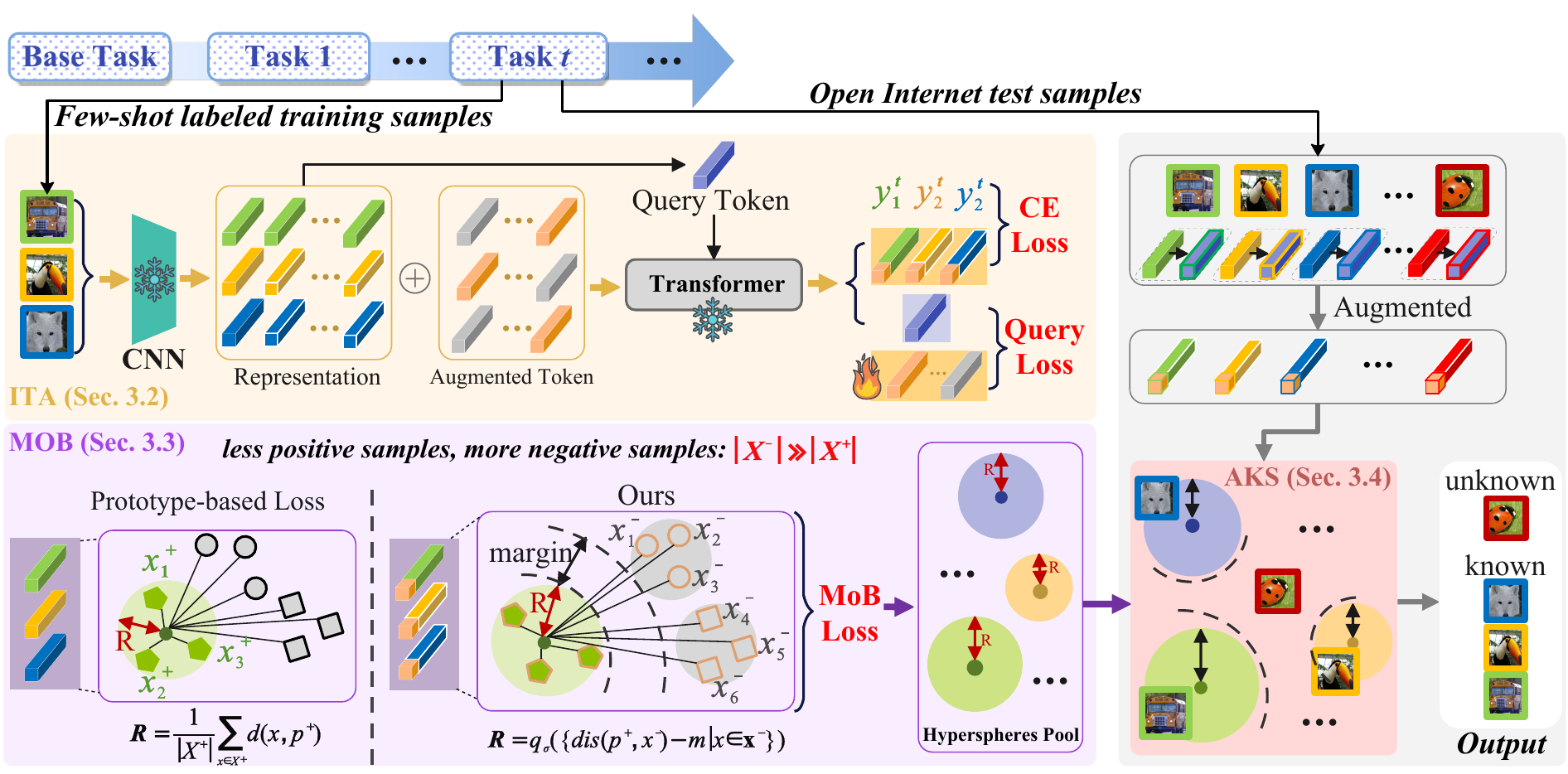}
\caption{The overall OFCL framework consists of three key components: (1) ITA (Yellow): enhancing the samples by matching them with appropriate additional tokens; (2) MOB (Purple): constructing compact decision boundaries of knowns by margin-based loss function for open detection; (3) AKS (Red): incorporating knowledge learned from both knowns and unknowns, and facilitating the transition from unknowns to knowns.}
\label{fig:framework}
\end{figure*}

\section{Methodology}
In this section, we begin by formulating the OFCL problem. Then, we detail the \textit{instance-wise token augmentation} (ITA) for knowledge transfer and present \textit{margin-based open boundary} (MOB). Lastly, we introduce an \textit{adaptive knowledge base} (AKB) designed to represent knowledge from unknowns and to facilitate updating unknowns into knowns in a hypersphere-based embedding space.

\textbf{\textit{Remarks.}}
Different from existing works, this paper addresses the limitations of OWCL by focusing on a more practical and challenging few-shot learning scenario, introducing an innovative concept of OFCL.
By integrating previous prompt-based learning strategies, our model can effectively adapt to limited labeled samples by ITA.
Moreover, MOB leverages hyperspherical embeddings to construct margin-based boundaries adaptively, ensuring compact and robust representations for open detection. AKS, on the other hand, pioneers an adaptive way to knowledge accumulation, transfer and replay, facilitating efficient knowledge utilization for both known and unknown samples, overcoming the limitations in previous OWCL methods (e.g., logits-based thresholds in Pro-KT \cite{li2024learning}).

\subsection{Problem Statement}
Our investigated problem, Open-world few-shot continual learning (OFCL), aims to incrementally learn within an open-world assumption where unseen or open samples may appear in the test phase, and generalize well on new tasks with scarce labeled training data.
It is important to note that we assume only the base task to have ample training samples (as shown in \autoref{fig:intro}), while each subsequent new task ($t > 0$) only involves scarce labeled training samples. 

\textbf{\textit{Problem Definition.}}
Given tasks $\{1, ..., t, ...\}$, each training set can be organized as $N$-way $K$-shot format $D^t_{tr} = \{ (x^t,C) \mid x^t \in X^t_{tr}(C), C\in CLS_{tr}^t \}$, where $X^t_{tr}(C)$ is a set of $K$-samples of the class $C$, $CLS_{tr}^t$ is the set of $N$ classes in the current task $t$. 
Similarly, the test set of task $t$ can be defined as $D^t_{te} = \{ (x^t,C) \mid x^t \in X^t_{te}(C), C \in CLS_{te}^t \}$. 
Here, due to the open-world assumption, we do not assume any `relations' between the classes in the test set and the training set. 
Specifically, some test samples may be unknown/open, i.e., belonging to the set $(CLS_{te}^t - CLS_{tr}^t)$. In testing, we evaluate on all the test samples $D^1_{te} \cup ... \cup D^t_{te}$ continually. 
Hence, the purpose of OFCL is to develop a unified classifier $f_{\theta}$ with parameter $\theta$  that (1) \textit{detects unknown/open samples}, (2) \textit{classifies known samples correctly}, and (3) \textit{supports the transforming of unknowns into knowns incrementally}.

\subsection{Instance-wise Token Augmentation (ITA)}

Inspired by the strong few-shot learning ability of prompt tuning \cite{qin2021lfpt5,wang2022learning,wang2022dualprompt} methods, we propose a knowledge augmentation paradigm for OFCL, referred to as Instance-wise Token Augmentation (ITA).
Notably, ITA can facilitate the representation, accumulation, and transfer of knowledge across various tasks to improve the knowledge transfer.

Given a task $t$ with its training set $D^t_{tr}$, the learner first initializes a batch of random tokens $\mathbf{p}^t$ as:
\begin{equation}
\mathbf{p}^t=(p_1^t,...,p_i^t,..., p_l^t), \text{with } p_i^t \in \mathbb{R}^{L_p \times D_e} \,.
\end{equation}
where $l$ is the number of additional tokens, $L_P$ is the token length and $D_e$ is the embedding dimension. 
During the training of task $t$, we maintain a token frequency set $\nu^t=(\nu_1,...,\nu_k,...)$,  where $\nu_k$ represents the frequency of each $p^t_k$ is selected to be appended into sample embeddings.

Hence, the essence of ITA lies in identifying and matching important tokens for sample augmentations. Here, we design an instance-wise mechanism to select useful additional tokens 
by looking up the top-$K$ keys:
\begin{equation}\label{equa:prompt}
(\mathbf{k}^*,\mathbf{p}^*)=\underset{K}{\operatorname{argmin}} \sum^l_{i=1} sim(\mathbf{h}^t, k^t_{i}) \cdot \nu_{i},
    \text{with } \mathbf{h}^t = \mathbf{Q}(x^t),
\end{equation}
where $k^t_i$ is the key associated with token $p^t_i$, the $sim(\cdot,\cdot)$ is a cosine similarity function and the $\mathbf{Q}$ is a query function encoding an input $x^t$ to the same dimension as the keys.
Next, the input embedding $\mathbf{h}^t$ is augmented by the above subset $\mathbf{p}^* \subset \mathbf{p^t}$:
\begin{equation}\label{equa:extend}
 \mathbf{h'}^t = \mathbf{h}^t \oplus  \mathbf{p}^*,
\end{equation}
where $\oplus$ denotes a dimension-wise concatenation. 

Subsequently, the improved loss function can be formulated as:
\begin{equation}\label{equa:prompt_loss}
    \begin{split}
        \mathcal{L}_{aug}= \mathcal{L}( f_{\theta}(f_{pr}(\mathbf{h'}^t),\mathbf{y}^t)+
        \lambda \cdot \sum_{i=1}^{l} sim(\mathbf{h}^t, k^{t}_{i}) ,
    \end{split}
\end{equation}
where $\mathbf{h}^t$ is the set of training samples projected embeddings, $\mathbf{y}^t$ are the training labels, $\lambda$ is a trade-off parameter, $f_{pr}$ is the pre-trained backbone. 
The first addend is the classification loss, and the second one is a surrogate loss to pull selected keys closer to corresponding query features.

Different from previous prompt-based CL methods, our proposed ITA stores the tokens learned from each task into a unified token bank $(\mathbf{K},\mathbf{P})$, treating them as knowledge to avoid forgetting. During testing, the model selects appropriate tokens from the token bank $(\mathbf{K},\mathbf{P})$ for each test sample. Hence, the proposed ITA not only enables the model to leverage prior knowledge but also mitigates the issue of overfitting with scarce labeled samples. 

\subsection{Margin-based Open Boundary (MOB)} 

By leveraging the augmented training samples, we facilitate the formation of more precise and compact decision boundaries for open sample detection. In contrast to existing approaches, our strategy goes beyond simply merging current continual learning techniques with open-set recognition. 

Thus, we introduce the MOB in a hypersphere space to construct compact boundaries from known samples or training classes, thereby naturally improving the open detection for OFCL. 
Hyperspheres outperform Euclidean distance in open-set recognition by reducing the \textit{hubness} problem \cite{trosten2023hubs}, where certain points disproportionately dominate nearest-neighbor lists, leading to misclassifications. 
In OFCL, where new, unseen classes may appear, the ability of hyperspherical embeddings to adapt is critical. These embeddings can continuously evolve as new data points are added, unlike Euclidean embeddings which can become skewed by hubs. Hyperspherical embeddings maintain robustness in dynamic, low-sample environments, improving their suitability for open-set recognition.

Given a class $C \in CLS_{tr}^t$, we divide all samples into a positive set  $\mathbf{x}^+ = \{ x^t | (x^t,C) \in D_{tr}^t \}$ containing all sample of the class $C$ at task $t$, and a negative set 
$\mathbf{x}^- = \{ x^t | (x^t,C') \in D_{tr}^t, C' \not= C \}$ containing all other samples of the same task. Our intuition here comes from contrastive learning, where training samples belonging to $C$ serve as positive samples, while all other training samples are treated as negative instances (in \autoref{fig:framework}).
Hence, the MOB loss function $\mathcal{L}_{margin}$ for constructing decision boundaries of knowns can be formulated as:
\begin{equation}\label{eq:loss_margin}
\begin{split}
    \mathcal{L}_{margin} & = \frac{1}{N} \cdot \sum_{C \in CLS{_{tr}}} 
    (\lambda \cdot r^2_{C}\\
   & +\frac{1}{\alpha} \cdot log[1+\sum_{x \in \mathbf{x}^+}{e^{\alpha  \cdot  (dis(\mathbf{c}_{C},f_{pr}(x))-r_{C})}}]\\
   & +\frac{1}{\beta} \cdot log[1+\sum_{x \in \mathbf{x}^- }e^{-\beta  \cdot  (dis(\mathbf{c}_C,f_{pr}(x))-(r_C+m))}]),
\end{split}
\end{equation}
where $N$ is the number of classes in the training set and $m$ is the margin, $r_{C}$ denotes the hypersphere radius of class $C$ around the centroid $\mathbf{c}_{C}$ that we discuss below, $dis(\cdot,\cdot)$ is a distance function, $\lambda$ is a constant that balances the trade-off between the regulation of radius and margin loss item, and $\alpha$ and $\beta$ are scaling factors for positive and negative sets, respectively. Note that the pre-trained backbone function $f_{pr}$ here takes as input a sample instead of the embedding of an input as in \autoref{equa:prompt_loss}.

Specifically, we employ three key components in the MOB: 
(1)~\textit{known classes centroids}: learning a centroid as the center of a hypersphere for each known class; 
(2)~\textit{margin and radius}: enforcing a margin between different classes while automatically learning a radius for each class;
(3)~\textit{open boundary}: identifying unknowns relying on the establishment of compact boundaries for known classes.

\textbf{Known Classes Centroids.} 
Hyperspherical embeddings distribute data more uniformly, avoiding central clustering, and methods like noHub \cite{trosten2023hubs} preserve class structure while maintaining uniformity. This balance allows hyperspheres to continuously adapt in dynamic, low-sample environments, improving accuracy in open-set scenarios.
Hence, we define an embedding space full of hyperspheres where points cluster around a single centroid representation for each class.

Given a class $C$ from task $t$, its centroid $\mathbf{c}_{C}$ is:
\begin{equation}\label{eq:centroid}
 {\mathbf{c}_C=\frac{1}{|\mathbf{x}^+|} \sum_{x^t \in \mathbf{x^+}} f_{pr}(x^t)}.
\end{equation}

Notably, during the training process of $\{ 1, ..., t,...\}$, the model stores all learned centroids incrementally. Each centroid is treated as the center of a hypersphere, and next, an appropriate radius needs to be determined for establishing compact boundaries between knowns and unknowns.

\textbf{Margin and Radius.}
Given a class $C$ with few-shot training samples, the number of negative samples $|\mathbf{x}^-|$ is typically much larger than the number of positive samples $|\mathbf{x}^+|$. 
To preserve distinct distributions for different classes, we constrain the distance between the class centroid $\mathbf{c}_{C}$ and every positive sample $x \in \mathbf{x}^+$ to be less than a learnable radius: 
\begin{equation}
{dis(\mathbf{c}_{C},f_{pr}(x))< r_{C}},
\end{equation}
where the radius $r_C$ is optimized by \autoref{eq:loss_margin}.
Accordingly, for every negative sample $x \in \mathbf{x}^-$ of class $C$:
\begin{equation}\label{eq:dis_neg2centroid}
{dis(\mathbf{c}_{C},f_{pr}(x))> r_{C}+m},
\end{equation}
where $m$ is the margin. 

Hence, the hypersphere of class $C$ is constructed by optimizing the radius $r_{C}$ as:
\begin{equation}\label{get_radius}
r_{C}=q_\sigma(\{dis(\mathbf{c}_{C},f_{pr}(x)) -m \mid  x \in \mathbf{x}^- \}),
\end{equation}
where $q_\sigma(\cdot)$ is a quantile function and $\sigma$ serves as the constraint governing deviations. 
Each hypersphere learned from task $t$ is formulated as: 
\begin{equation}
(CLS^t_{tr},\mathbf{C}^{t},\mathbf{R}^{t}) = \{(C^t_1,\mathbf{c}_{C^t_1},r_{C^t_1}),...,(C^t_N,\mathbf{c}_{C^t_N},r_{C^t_N})\},
\end{equation}
where $N$ is the number of classes in $CLS^t_{tr}$.

During the training process of $\{ 1, ..., t,...\}$, the model stores all learned hyperspheres together:
\begin{equation}
\!\!\!(CLS_{tr},\mathbf{C},\mathbf{R}) = \{(CLS^1_{tr},\mathbf{C}^{1},\mathbf{R}^{1}),...,
(CLS^t_{tr},\mathbf{C}^{t},\mathbf{R}^{t}),... \}.
\end{equation}

\textbf{Open Detection.} 
In testing phrases, given a test sample $x$ from an arbitrary task, we initially project $x$ into all learned hyperspheres. Subsequently, to determine whether the test sample $x$ is unknown, we check the inclusion of $x$ in all hyperspheres:
If there is a class $C$ such that the sample is contained within the hypersphere of $C$, then $x$ is classified within $C$, otherwise, $x$ is identified as unknown. 

Then, the open detection procedure at task $t$ is:

\textbf{Step 1: }Identify the centroid $\mathbf{c}_C^*$ closest to $x$;

\textbf{Step 2: }Calculate the distance between the selected centroid $\mathbf{c}_C^*$ and the test sample $x$;

\textbf{Step 3: }If $d^* \leq r^*$, then $x$ belongs to class $C$ otherwise it is classified as unknown.

In MOB, we treat the hypersphere centers as learnable parameters, dynamically updating them based on deep feature representations during continual learning. Our method shares similarities with deep distance metric classifiers that employ margin-based loss functions. Moreover, the prototypes are set to hypersphere centers, which are adaptively updated throughout the learning process, i.e., in the adaptive knowledge space (AKS).

\subsection{Adaptive Knowledge Space (AKS)}
Considering $(CLS_{tr},\mathbf{C},\mathbf{R})$ as the knowledge learned from knowns, we present an Adaptive Knowledge Space (AKS), designed not only to store and transfer knowledge about unknowns acquired from previously learned tasks, but also to facilitate updating unknown knowledge into known knowledge during OFCL. 

\paragraph{Unknown Knowledge Representation.} 
After identifying unknowns, we then employ clustering on the identified unknown samples as follows:
Given a detected unknown sample $x$ let $\Gamma_\epsilon(x)$ represent the $\epsilon$-neighborhood of $x$, $MinPts$ denotes the minimum number of objects in $\Gamma_\epsilon(x)$, and $\rho(x)=|\Gamma_\epsilon(x)|$ be its density value. We define the boolean function $T(x) = 1$ if 
$\rho(x_j) \ge MinPts$ and $T(x) = 0$ otherwise.
If $T(x) = 1$, the unknown sample $x$ is clustered with density-reachable points, and the entire space is partitioned into $M$ groups $\{G_1, G_2, ..., G_M\}$. Otherwise, $x$ is treated as noise. 
Each group is then assigned a group centroid, which is determined using \autoref{eq:centroid}. 
The corresponding radius is obtained by \autoref{get_radius}.

Since there are no \textit{ground-truth labels} for unknowns, we generate a set of pseudo-labels and assign them to each group. Similar to hyperspheres learned from known samples, we have :
\begin{equation}
\!\!\!(CLS_{open}^t,\overline{\mathbf{C}^{t}},\overline{\mathbf{R}^{t}}) = \{(\overline{C^t_1},\mathbf{c}_{\overline{C^t_1}},r_{\overline{C^t_1}}),..., (\overline{C^t_M},\mathbf{c}_{\overline{C^t_M}},r_{\overline{C^t_M}})\}.
\end{equation}

\paragraph{Updating Unknowns to Knowns.}
At task $t$ the integrated adaptive knowledge space consists of all the learned hyperspheres during training $(CLS_{tr},\mathbf{C},\mathbf{R})$ together with all the new $(CLS_{open}^t,\overline{\mathbf{C}^{t}},\overline{\mathbf{R}^{t}})$ associated with the unknowns at all task until $t$. Because of this incremental growth, if previously encountered unknowns reappear, they are classified with corresponding pseudo-labels. However, if a new hypersphere learned during a subsequent task's training overlaps with an existing hypersphere with a pseudo-label, then we integrate the corresponding pseudo-category into the newly trained categories, thus facilitating a transition from unknowns to knowns.

\subsection{Overall Objective}
At any task $t$, OFCL first obtains $L$-additional tokens via $\mathcal{L}_{aug}$ and stores all the newly learned additional tokens together with those previously learned. Subsequently, each sample is augmented by additional knowledge, and an adaptive knowledge space containing knowledge learned from knowns and unknowns is constructed. 
Hence, the overall objective is:
\begin{equation}\label{equa:overall_loss}
\underset{(\mathbf{C}^{t},\mathbf{R}^{t}),\theta,(\mathbf{k}^{t},\mathbf{p}^{t})}{\operatorname{min}} \gamma \cdot \mathcal{L}_{margin}+(1-\gamma) \cdot \mathcal{L}_{aug},
\end{equation}
where the $\gamma$ is a balance between $\mathcal{L}_{margin}$ and $\mathcal{L}_{agu}$.

\begin{algorithm}[ht]
\caption{\textbf{: Training Process of the OFCL framework}}\label{algorithm1}
\justifying \textbf{Input}: the training set $D_{tr}^t=\{(\mathbf{X}^t,\mathbf{Y}^t)\}$, the set of training classes $CLS_{tr}^t$, the token bank $(\mathbf{K},\mathbf{P})=\{(\mathbf{k}^{t},\mathbf{p}^{t})\}_{t=1}^{T}$, a trainable classifier $f_{\theta}$, a backbone $f_{pr}$ and the hyperparameter balance $\gamma$.\\
\textbf{Initialize}: $(\mathbf{K},\mathbf{P})$, $f_\theta$.
\begin{algorithmic}[1] 
\FOR{t = 1, 2, ..., $T$}
\FOR{each epoch}
\FOR{$(\mathbf{x}^t,\mathbf{y}^t)$ in each training set $D_t^{tr}$ with class $CLS^t_{tr}$}
\STATE Map ${\mathbf{x}^n}$ to ${\mathbf{h}^t}$;
\STATE Get the similarity of ${h^t}$ and each key of $(\mathbf{k}^{t},\mathbf{p}^{t})$;
\STATE Select augmentation tokens with $(\mathbf{k}^{*},\mathbf{p}^{*})$ \autoref{equa:prompt};
\STATE Enhance ${\mathbf{h}^t}$ to $\mathbf{h'}^t$ with $\mathbf{p}^{*}$ via \autoref{equa:extend};
\STATE Calculate the augmented feature by $f_{pr}(\mathbf{h'}^t)$;
\STATE Calculate $\mathbf{c}_{C}$ for each class $C$ in $CLS^t_{tr}$ via \autoref{eq:centroid};
\STATE Obtain positive and negative set of class $C: \mathbf{x}^+, \mathbf{x}^-$;
\STATE Obtain margin and radius via \autoref{get_radius}
\STATE Calculate the overall loss via \autoref{equa:overall_loss};
\ENDFOR
\STATE Update $f_\theta$, $(\mathbf{K},\mathbf{P})$, the radius $r_C$ and margin $m$ by backpropagation;
\ENDFOR
\STATE Update the overall AKS with \textbf{1)} Hyperspheres learned from $D^t_{tr}$: $(CLS^t_{tr},\mathbf{C}^{t},\mathbf{R}^{t})$ and \textbf{2)} the Hyperspheres learned from unknowns: $(CLS_{open}^t,\overline{\mathbf{C}^{t}},\overline{\mathbf{R}^{t}})$.
\ENDFOR
\end{algorithmic}
\end{algorithm}

To better illustrate the proposed OFCL framework, we detail the training process in Algorithm \autoref{algorithm1}. 
First, the token bank and classifier are initialized. Then, for each time step \( t \), multiple training epochs are conducted, where each sample \( (\mathbf{x}^t,\mathbf{y}^t) \) undergoes feature mapping, similarity computation with stored keys, augmentation via token selection, and enhancement through the backbone network. Class-wise centroids, positive and negative sets, margins, and radii are computed to derive the overall loss, which is minimized via backpropagation. Finally, the AKS is updated with hyperspheres representing both known and unknown knowledge, enabling continual adaptation to evolving data distributions.

Accordingly, after learning task t, the model follows four steps during inference:

\textbf{Step 1}: Integrate new task knowledge (i.e., knowledge from labeled samples) into the current knowledge space, updating unknowns to knowns once the labels for open samples are learned.

\textbf{Step 2}: During testing, detect open samples and ensure they are not misclassified as known categories.

\textbf{Step 3}: For samples identified as knowns, classify them correctly while avoiding forgetting of previous tasks.

\textbf{Step 4}: Update the knowledge space adaptively by incorporating the new knowledge from the open sample into the current knowledge space.

In summary, our OFCL framework not only enhances the representation of known classes but also improves decision boundary compactness for open detection. This integration demonstrates our innovative development of existing techniques to address the unique challenges of OFCL.

\section{Experiments and Results}
\subsection{Experimental Setup}
\paragraph{Datasets.}
We follow the dataset configuration in TOPIC \cite{tao2020few}:
(1) The \textit{CUB200 dataset} \cite{wah2011caltech} encompasses 11,788 images depicting 200 distinct bird species. We partition the 200 classes into 100 base classes for task $0$ and 100 incremental classes for 10 tasks. Each task consists of 10 classes, with each class having 5 training samples (i.e., a 10-way 5-shot setup).
(2) The \textit{MiniImageNet dataset} \cite{vinyals2016matching} comprises 100 classes with a total of 60,000 RGB images. We partition the 100 classes into 60 base classes for an initial base task and 40 incremental classes for 8 tasks. Each task consists of 5 classes, each with 5 training samples (i.e., a 5-way 5-shot configuration).

\paragraph{Implementation Details.}

(1) We randomly shuffled the order of tasks in experiments. All results are the averages obtained from three random shuffles.
(2) Due to the current absence of available comparative baselines for OFCL, we divided our experiments into two parts for clear and fair comparisons: unknown detection and known classifications.
(3) All baselines are trained by the Adam optimizer with a batch size of 25 and a learning rate of 0.03. 

\paragraph{Baselines.}
To ensure comprehensive and fair comparisons, we consider two parts in the evaluation:

\textbf{Unknown Detection}: we compare our \textit{OFCL} against 8 competitive open detection baselines: 
\emph{MSP} \cite{hendrycks2016baseline}, 
\emph{KL} \cite{hendrycks2022scaling},
\emph{SSD} \cite{sehwag2021ssd},
\emph{MaxLogit} \cite{basart2022scaling},
\emph{Energy} \cite{liu2020energy}, 
\emph{ViM} \cite{wang2022vim},
\emph{KNN-based OOD} \cite{sun2022out},
\emph{NNGuide}  \cite{park2023nearest}.

\textbf{Known Classification}: we evaluate \textit{OFCL} against 15 benchmark continual learning methods with scarce labeled data: \textit{iCaRL} \cite{rebuffi2017icarl}, \textit{TOPIC} \cite{tao2020few}, \textit{CEC} \cite{zhang2021few},  \textit{Meta-FSCIL} \cite{chi2022metafscil}, \textit{ALICE} \cite{peng2022few}, \textit{FeSSSS} \cite{ahmad2022few}, \textit{Pro-KT} \cite{li2024learning}, \textit{LIMIT} \cite{zhou2022few}, \textit{NC-FSCIL} \cite{yang2022neural}, \textit{MCNet} \cite{ji2023memorizing}, \textit{SoftNet} \cite{yoon2023soft}, \textit{CoSR} \cite{wang2023continual}, \textit{M-FSCIL} \cite{li2022memory}, \textit{FSIL-GAN} \cite{agarwal2022semantics}, and \textit{CPE-CLIP} \cite{d2023multimodal}.


\paragraph{Metrics.}
For unknown detection, we use the average area under the ROC across all learned $N$-tasks $AUC_N$ and the average false positive rate across $N$-tasks $FPR_N$ as metrics \cite{chan2021entropy}. 
For known classification, we apply averaged accuracy $ACC_N$ as the metric, where $ACC_N$ is calculated over $N$ new tasks.
To measure forgetting, we calculate the performance dropping rate PD = $ACC_0-ACC_N$, where $ACC_0$ is the classification accuracy in the base task \cite{li2024learning}.

\begin{table}[ht]
\centering\small
\begin{tabular}{c|c|cc}
\hline
 Dataset & Methods &$AUC_N$ $\uparrow$ &$FPR_N$ $\downarrow$ \\ \hline
\multicolumn{1}{c|}{\multirow{8}{*}{\rotatebox{90}{\begin{tabular}[c]{@{}c@{}}CUB200\\ (10-way 5-shot)\end{tabular}}}} 
& MSP & 61.34 & 92.95\\
\multicolumn{1}{c|}{} & KL & 60.81 & 93.10 \\
\multicolumn{1}{c|}{} & SSD & 37.70 & 99.52 \\
\multicolumn{1}{c|}{} & MaxLogits & \underline{61.31} & \underline{92.43} \\
\multicolumn{1}{c|}{} & Energy & 60.81 & 93.10\\
\multicolumn{1}{c|}{} & ViM & 54.16 & 97.91 \\
\multicolumn{1}{c|}{} & KNN-based OOD & 37.20 & 99.16 \\
\multicolumn{1}{c|}{} & NNGuide & 50.26 & 96.97 \\
 \cline{2-4} 
\multicolumn{1}{c|}{} & OFCL & \textbf{69.72} & \textbf{69.96}  \\ 
\hline
\multirow{8}{*}{\rotatebox{90}{\begin{tabular}[c]{@{}c@{}}MiniImageNet\\ (5-way 5-shot)\end{tabular}}} 
 & MSP & 49.78 & 94.91 \\
\multicolumn{1}{c|}{} & KL & 46.61 & 95.10 \\
\multicolumn{1}{c|}{} & SSD & 50.64 & 95.18 \\
\multicolumn{1}{c|}{} & MaxLogits & 46.78 & 94.65 \\
\multicolumn{1}{c|}{} & Energy & 46.61 & 95.10\\
\multicolumn{1}{c|}{} & ViM & \underline{59.56} & \underline{89.87} \\
\multicolumn{1}{c|}{} & KNN-based OOD & 52.88 & 93.78 \\
\multicolumn{1}{c|}{} & NNGuide & 52.47 & 90.95 \\
 \cline{2-4} 
 & OFCL & \textbf{76.20} & \textbf{71.30}\\ 
 \hline 
\end{tabular}
\caption{Results($\%$) regarding unknown detection. We report the results over 10 tasks for CUB200 (10-way 5-shot) and 8 tasks for MiniImageNet (5-way 5-shot).}
\label{table:open detection}
\vspace{-0.8cm}
\end{table}

\begin{table*}[ht]
\centering\small
\begin{tabular}{l@{\hspace{8pt}}c@{\hspace{8pt}}c@{\hspace{8pt}}c@{\hspace{8pt}}c@{\hspace{8pt}}c@{\hspace{8pt}}c@{\hspace{8pt}}c@{\hspace{8pt}}c@{\hspace{8pt}}c@{\hspace{8pt}}c@{\hspace{8pt}}c@{\hspace{8pt}}c@{\hspace{8pt}}c@{\hspace{8pt}}}
\hline
\multicolumn{1}{c}{\multirow{2}{*}{Methods}}  & \multicolumn{11}{c}{$ACC_N$ in each task (\%) $\uparrow$}  & \multirow{2}{*}{PD$\downarrow$} \\ \cline{2-12}
\multicolumn{1}{c}{}        & 0 & 1 & 2 & 3 & 4 & 5 & 6 & 7 & 8 & 9 & 10 &  &  \\ \hline
                iCaRL      & 68.68 & 52.65 & 48.61 & 44.16 & 36.62 & 29.52 & 27.83 & 26.26 & 24.01 & 23.89 & 21.16 & 47.52  \\
                {TOPIC}       & 68.68 & 62.49 & 54.81 & 49.99 & 45.25 & 41.40 & 38.35 & 35.36 & 32.22 & 28.31 & 26.28 & 42.40  \\
                CEC         & 75.85 & 71.94 & 68.50 & 63.50 & 62.43 & 58.27 & 57.73 & 55.81 & 54.83 & 53.52 & 52.28 & 23.57  \\
                Meta-FC  & 75.90 & 72.41 & 68.78 & 64.78 & 62.96 & 59.99 & 58.30 & 56.85 & 54.78 & 53.82 & 52.64 & 23.26  \\     
               ALICE        & 77.40 & 72.70 & 70.60 & 67.20 & 65.90 & 63.40 & 62.90 & 61.90 & 60.50 &  60.60 &  60.10 &  17.30  \\    
                FeSSSS      & 79.60 & 73.46 & 70.32 & 66.38 & 63.97 & 59.63 & 58.19 & 57.56 & 55.01 & 54.31 & 52.98 & 26.62  \\
                Pro-KT & \textbf{82.90} & 74.15 & 72.82 & 62.46 & 59.69 & 54.88 & 51.56 & 48.57 & 47.32 & 46.86 & 47.68 &  35.22 \\
                LIMIT       & 76.32 & 74.18 & 72.68 & 69.19 &  68.79 &  65.64 & 63.57 &  62.69 &  61.47 & 60.44 & 58.45 & 17.87  \\   
                NC-FSCIL    & 80.45 & 75.98 &  72.30 &  70.28 & 68.17 & 65.16 &  64.43 & 63.25 & 60.66 & 60.01 & 59.44 & 21.01  \\
                MCNet       & 77.57 & 73.96 & 70.47 & 65.81 & 66.16 & 63.81 & 62.09 &  61.82 & 60.41 & 60.09 & 59.08 & 18.49 \\  
                SoftNet     & 78.07 & 74.58 & 71.37 & 67.54 & 65.37 & 62.60 & 61.07 & 59.37 & 57.53 & 57.21 & 56.75 & 21.32  \\      
                CoSR        & 74.87 & 73.15 & 68.23 & 63.50 & 62.72 & 59.10 & 57.46 & 55.73 & 53.28 & 52.31 & 51.75 & 23.12 \\ 
               M-FSCIL &81.04 &\underline{79.73} &76.62&73.30&71.22&68.90&66.87&65.02&63.90&62.49&60.40&20.64\\
               FSIL-GAN & 81.27 & 78.03& 75.61& \underline{73.72} & 70.49& 68.19& 66.58& 65.63& 63.21& 60.92& 59.43&21.84\\
               CPE-CLIP & \underline{81.58}& 78.52& \underline{76.68} & 71.86& \underline{71.52} & \underline{70.23}& \underline{67.66} & \underline{66.52} & \underline{65.09} & \underline{64.47} & \underline{64.60} & \underline{16.98} \\
                \hline
                OFCL          & 76.67 & \textbf{81.09} & \textbf{76.80} & \textbf{76.20} & \textbf{74.80} & \textbf{72.44} & \textbf{71.52} & \textbf{69.17} & \textbf{70.33} & \textbf{68.17} & \textbf{68.92} & \textbf{7.74} \\
                 \hline
\end{tabular}
\caption{Averaged classification accuracy (\%) on CUB200 (10-way 5-shot).}
\vspace{-3mm}
\label{tab:known-cub}
\end{table*}

\begin{table*}[ht]
\centering
\begin{tabular}
{l@{\hspace{8pt}}c@{\hspace{8pt}}c@{\hspace{8pt}}c@{\hspace{8pt}}c@{\hspace{8pt}}c@{\hspace{8pt}}c@{\hspace{8pt}}c@{\hspace{8pt}}c@{\hspace{8pt}}c@{\hspace{8pt}}c@{\hspace{8pt}}c@{\hspace{8pt}}}
\hline
\multicolumn{1}{c}{\multirow{2}{*}{Methods}}  & \multicolumn{9}{c}{$ACC_N$ in each task (\%) $\uparrow$}  & \multirow{2}{*}{PD $\downarrow$} \\ \cline{2-10}
\multicolumn{1}{c}{}        & 0 & 1 & 2 & 3 & 4 & 5 & 6 & 7 & 8 &  &  \\ \hline           
                iCaRL      & 61.31 & 46.32 & 42.94 & 37.63 & 30.49 & 24.00 & 20.89 & 18.80 & 17.21 & 44.10   \\     
                {TOPIC}    & 61.31 & 50.09 & 45.17 & 41.16 & 37.48 & 35.52 & 32.19 & 29.46 & 24.42 & 36.89 \\
                CEC        & 72.00 & 66.83 & 62.97 & 59.43 & 56.70 & 53.73 & 51.19 & 49.24 & 47.63 & 24.37 \\
                Meta-FSCIL      & 72.04 & 67.94 & 63.77 & 60.29 & 57.58 & 55.16 & 52.9 & 50.79 & 49.19 & 24.41  \\     
                        
                ALICE           & 80.60 & 70.60 & 67.40 & 64.50 & 62.50 & 60.00 & 57.80 & 56.80 & 55.70 & 24.90 \\          
                FeSSSS          & 81.50 & 77.04& 72.92 & 69.56 & 67.27 & 64.34 &  62.07 &  60.55 &  58.87 &  22.63 \\
                Pro-KT & \textbf{98.59} & 79.82 & 77.85 & 71.82 & 70.34 & 69.18 & 70.13 & 69.27 & 69.73 & 28.86 \\
                LIMIT           & 72.32 & 68.47 & 64.30 & 60.78 & 57.95 & 55.07 & 52.70 & 50.72 & 49.19 & 23.13 \\         
                NC-FSCIL        &  84.02 &  76.80 &  72.00 &  67.83 &  66.35 &  64.04 & 61.46 & 59.54 & 58.31 & 25.71 \\
                MCNet           & 72.33 & 67.70 & 63.50 & 60.34 & 57.59 & 54.70 &  52.13 & 50.41 & 49.08 & 23.25  \\  
                SoftNet         & 76.63 & 70.13 & 65.92 & 62.52 & 59.49 & 56.56 & 53.71 & 51.72 & 50.48 & 26.15  \\      
                CoSR            & 71.92 & 66.91 & 62.71 & 59.59 & 56.63 & 53.78 & 51.01 & 49.23 &  47.93 & 23.99 \\ 
                M-FSCIL & 93.45 & \textbf{91.82}& 87.09& \textbf{88.07}& \textbf{86.75}& \underline{87.15} & \underline{85.68} & \underline{84.80} & \underline{85.37} &\underline{8.08}\\
                FSIL-GAN &69.87& 62.91& 59.81& 58.86& 57.12& 54.07& 50.64& 48.14& 46.14&23.73\\
                CPE-CLIP &90.23& 89.56& \underline{87.42} & \underline{86.80}& \underline{86.51} & 85.08& 83.43& 83.38& 82.77&\textbf{7.46}\\
                \hline
                
                OFCL             & \underline{96.00} & \underline{91.65} & \textbf{89.17} & 84.90 & 86.46 & \textbf{87.18} & \textbf{87.57} & \textbf{85.71} & \textbf{87.12} & 10.88\\
                \hline
\end{tabular}
\caption{Averaged classification accuracy (\%) on MiniImageNet (5-way 5-shot).}
\label{tab:known-mini}
\vspace{-0.3cm}
\end{table*}

\paragraph{Results on Unknown Detection.} 
\autoref{table:open detection} presents the results of unknown detection. Notably, the baselines exhibit poor performance under the OFCL setting, underscoring their difficulty in adapting to real-world scenarios. 

In contrast, OFCL consistently outperforms all compared methods across various configurations, as indicated by its superior $AUC_N$ and $FPR_N$ values. This consistent advantage underscores the robustness and adaptability of OFCL in handling diverse learning conditions. The ability of OFCL to delineate precise decision boundaries and establish an adaptive knowledge space further highlights the effectiveness of our proposed approach in addressing complex knowledge boundaries. By dynamically integrating open-set detection with continual learning, OFCL effectively mitigates the challenges posed by evolving knowledge distributions. 
Overall, OFCL demonstrates significant improvements demonstrated by OFCL in open detection performance within real-world scenarios characterized by knowledge boundaries.

\begin{figure}[ht]
\centering
\includegraphics[width=\columnwidth]{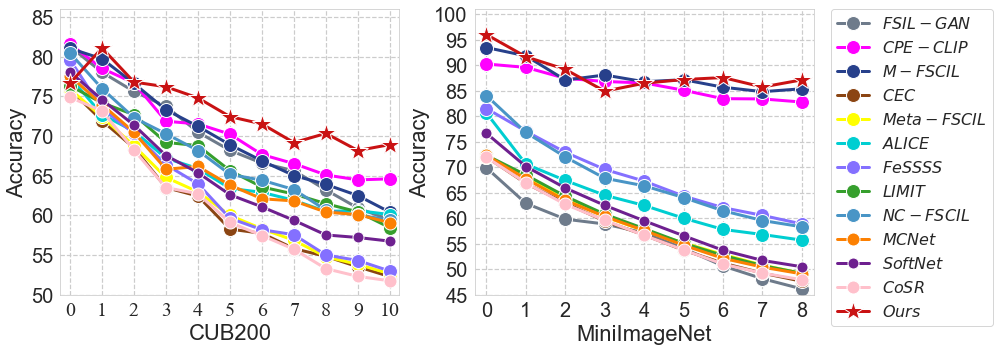}
\vspace{-5mm}
\caption{Visualizations of Known Classification.}\label{fig:baseline}
\end{figure}

\paragraph{Results on Known Classification.}
In this experiment, we evaluate the model's performance on known classification against various competitive benchmarks under a same setting. 
\autoref{tab:known-cub} and \autoref{tab:known-mini} present the accuracy scores achieved by different methods on the CUB200 and MiniImageNet datasets, respectively.

Despite the challenges of the OFCL problem, the proposed framework consistently outperforms other baselines, demonstrating lower drop rates and higher final accuracy across all classes. These results highlight the effectiveness of instance-wise token augmentations in mitigating catastrophic forgetting and overfitting. Furthermore, as shown in \autoref{fig:baseline}, the growing performance advantage of OFCL over time suggests superior adaptability in long-term learning scenarios compared to existing baselines.

Notable, on the MiniImageNet dataset, CPE-CLIP achieves a PD score of 7.46, outperforming ours. 
CPE-CLIP enhances CLIP's pre-training mechanism, improving performance across various cross-modal tasks (e.g., image retrieval, text generation). Its pre-training dataset overlaps with MiniImageNet, which may contribute to its competitive PD score.  
Additionally, in the base task, our model achieves an $ACC_N$ of 96, compared to 90 for CPE-CLIP, further contributing to the higher PD score. 
Hence, these results further underscore the robustness of our OFCL in CL scenarios.

\subsection{Ablation Studies}

We conduct ablation studies, analyzing the impact of three key components: 1) the additional tokens selection mechanism in ITA, 2) the whole instance-wise token augmentation, and 3) the whole AKS. The average final accuracy $ACC_N$ represents the mean accuracy over the past $N$ tasks, and results are presented in \autoref{tab:Ablation study}.

First, removing the additional token selection mechanism in ITA leads to a noticeable decline in both known classification and unknown detection performance. This highlights its critical role in mitigating overfitting at knowledge boundaries and enhancing open-set detection.  
Second, eliminating ITA results in the most significant degradation in known classification, underscoring its effectiveness in reducing overfitting and addressing CF.  
Finally, removing the entire AKS module completely disables open-set detection. As shown in \autoref{tab:Ablation study}, this also degrades knowns classification performance, indicating that the softmax mechanism alone is insufficient for distinguishing between known classes.  

Moreover, our method is highly decoupled from specific backbone architectures, as it operates at the input and output levels, ensuring broad adaptability. To validate this flexibility, we conducted an ablation study using ViT, a widely adopted large-scale image classification model, as the backbone with fine-tuning.  
On the CUB200 dataset, OFCL achieves a significantly higher known classification accuracy (\textbf{63.80\%}) compared to ViT with fine-tuning alone (\textbf{21.19\%}). This substantial performance gap demonstrates that strong representational capacity alone is insufficient, highlighting the effectiveness of our framework in facilitating structured knowledge transfer and open-set adaptation.

\begin{table}[ht]
\centering
\begin{tabular}{c@{\hspace{3pt}}c@{\hspace{3pt}}c@{\hspace{3pt}}c@{\hspace{3pt}}c@{\hspace{3pt}}}
\hline
\multirow{2}{*}{\textbf{\begin{tabular}[c]{@{}c@{}}Unknowns \\ Detection\end{tabular}}} & \multicolumn{2}{c}{CUB200} & \multicolumn{2}{c}{MiniImageNet} \\ \cline{2-5}
 & $AUC_N$ & $FPR_N $ & $AUC_N$  & $FPR_N$ \\ \hline
 w/o selection mechanism & 66.68 & 69.96 & 75.73 & 80.54 \\
w/o ITA & 67.06 & 75.83 & 70.61 & 86.71 \\
w/o AKS & - & - & - & - \\
OFCL   & \textbf{69.72} & \textbf{69.85} & \textbf{76.20} & \textbf{71.30} \\ \hline
\multirow{2}{*}{\textbf{\begin{tabular}[c]{@{}c@{}}Knowns \\ Classification\end{tabular}}} &\multicolumn{2}{c}{CUB200} & \multicolumn{2}{c}{MiniImageNet} \\ \cline{2-5}
 & $ACC_N$ & PD & $ACC_N$ & PD \\ \hline
w/o selection mechanism & 54.52 & 12.48 & 61.34 & 29.66 \\
w/o ITA & 50.33 & 31.14 & 51.42 & 34.58 \\
w/o AKS & 55.24 & 28.42 & 63.02 & 34.78 \\
OFCL & \textbf{63.80} & \textbf{8.20} & \textbf{65.24} & \textbf{25.56}\\ \hline
\end{tabular}
\caption{Ablation studies.}
\label{tab:Ablation study}
\vspace{-9mm}
\end{table}

\subsection{Additional Results}

\paragraph{Time and Space Complexity.}
As new tasks continually emerge, the complexity of AKS scales linearly with $\mathcal{O}(n)$. 
For the CUB200, after training all tasks, the final AKS only occupies 0.048 MB. 
As for the MiniImageNet, the final AKS requires 0.024 MB. 
The lookup time for instance-wise tokens matching is 0.001 seconds, and the decision time is less than 0.01 seconds. 
Moreover, we evaluated inference time: instance-wise token matching takes 0.001s, and classification is under 0.01s. Following your comment, we added experiments on: training one epoch took around 0.4284s (on CUB200) and 0.8279s (on MiniImageNet); peak memory usage is less than 28869632B, i.e., 28.87 M.

\begin{figure}[ht]\centering
   \subfloat[]{
    \includegraphics[width=0.185\textwidth]{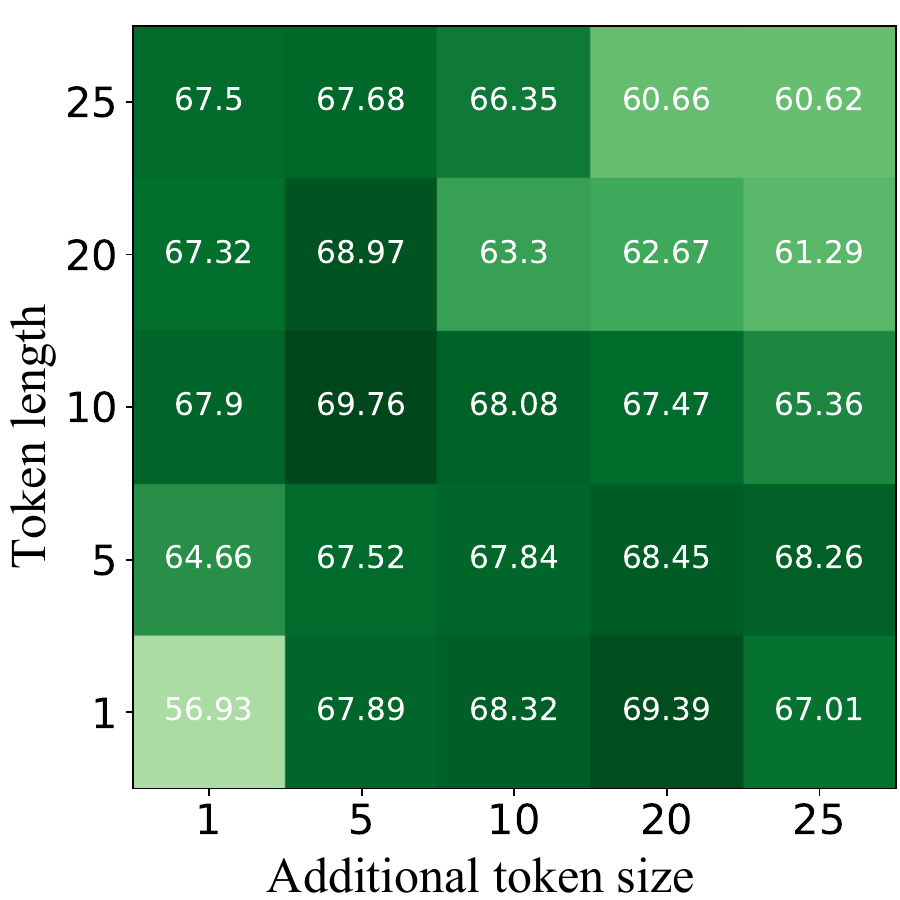}
    }
    \subfloat[]{
    \includegraphics[width=0.2\textwidth]{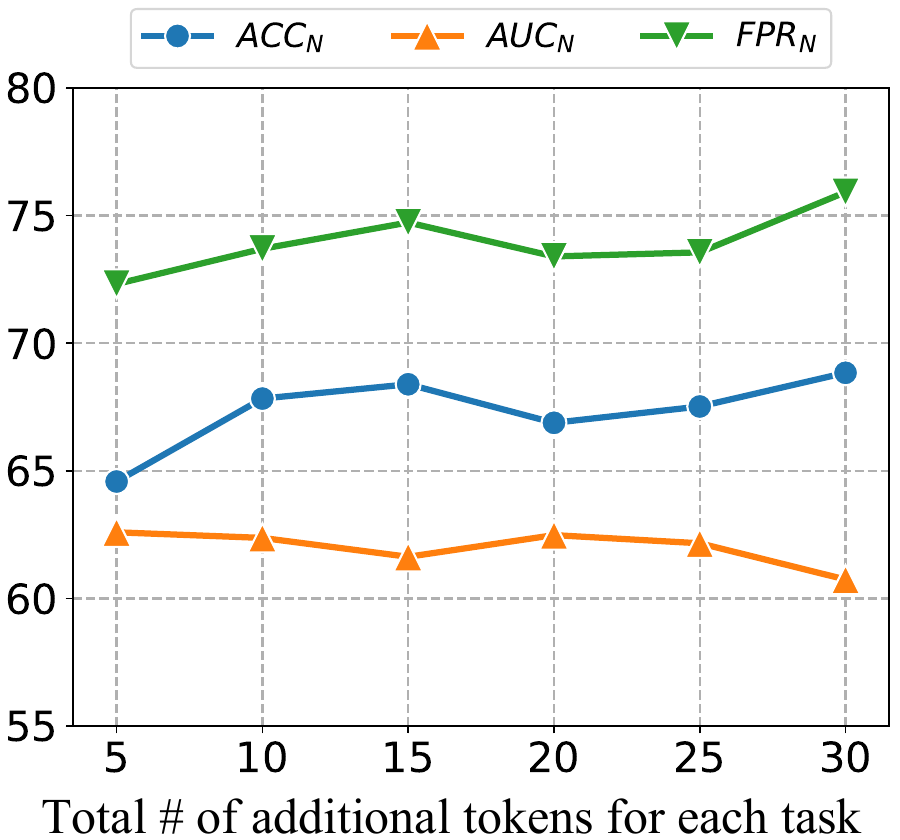}
    }
    \vspace{-3mm}
    \caption{(a): $ACC_N$ w.r.t token length $L_P$ and additional token size $K$, given $l=25$. (b): $ACC_N$ w.r.t. $l$ (i.e., the total number of additional tokens of each task) with $L_P=5$ and $K=5$ (take CUB200 for illustration).}
    \label{fig:param_prompt}
\vspace{-0.5cm}
\end{figure}

\begin{figure}[ht]\centering
    \subfloat[]{
    \includegraphics[width=0.2\textwidth]{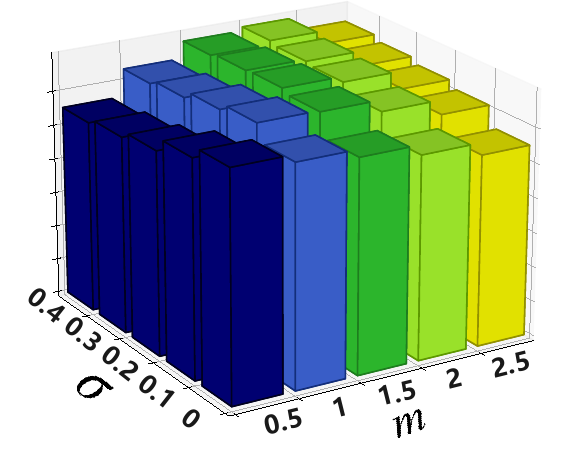}
    }
    \subfloat[]{
    \includegraphics[width=0.2\textwidth]{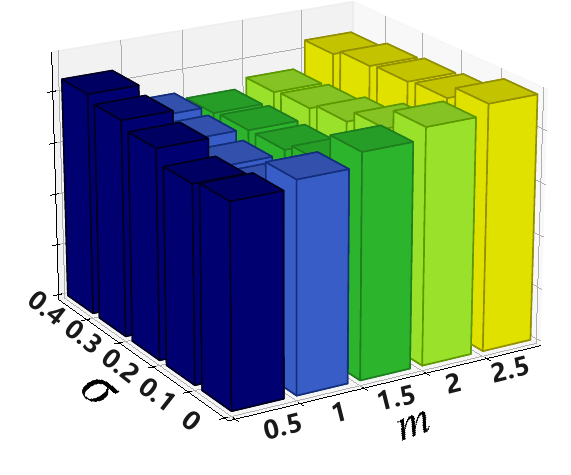}
    }
    \vspace{-3mm}
    \caption{(a) and (b): $AUC_N$ and $FPR_N$ w.r.t constraint governing deviations $\sigma$ and margin $m$ (CUB200 dataset).}
    \label{fig:3D}
\vspace{-5mm}
\end{figure}

\paragraph{Parameter Sensitivity Analysis.}
Consider the crucial hyperparameters, namely (1) the margin $m$ and the constraint governing deviations $\sigma$ (in \autoref{get_radius}), and (2) the token length $L_P$, additional token size $K$, and the amount of additional tokens per task $l$. 

\autoref{fig:param_prompt} (a) shows that using a large selection size $K$ and a long token length $L_P$ may lead to knowledge under-fitting, whereas \autoref{fig:param_prompt} (b) demonstrates that the impact of $l$ remains relatively stable within the range $5 \le l \le 30$.
The parameter $\sigma$ acts as a constraint, governing deviations and regulating boundary violations. A smaller $\sigma$ imposes a more compact boundary. As depicted in \autoref{fig:3D}, when $m=0.5$, performance decreases as $\sigma$ increases. However, for larger margins, performance improves with increasing $\sigma$, indicating the robustness of open detection is sensitive to the margin $m$. 
Additionally, another key hyperparameter in our framework is $\gamma$ (as defined in \autoref{equa:overall_loss}), which controls the trade-off among three components in the overall loss: classification loss for known samples, similarity measure in ITA, and distance measure in MOB, i.e., (classification loss : $sim(\cdot)$ : $dis(\cdot)$).
As shown in the \autoref{tab:gamma}, the performance of our model remains consistently stable across a wide range of $\gamma$ configurations. This demonstrates the strong robustness of the proposed OFCL framework with respect to the loss weight settings.

\begin{table}[h]
\centering
\begin{tabular}{lccc}
\toprule
\textbf{$\gamma$} & \textbf{$ACC_N$} & \textbf{$AUC_N$} & \textbf{$FPR_N$} \\
\midrule
1:00:01       & 57.18  & 62.12  & 73.81 \\
1:0.1:0.9     & 57.78  & 62.37  & 72.85 \\
1:0.3:0.7     & 57.77  & 62.47  & 73.25 \\
1:0.5:0.5     & 58.05  & 62.46  & 73.35 \\
1:0.7:0.3     & 57.42  & 62.50  & 72.30 \\
1:0.9:0.1     & 57.53  & 62.08  & 73.75 \\
1:01:00       & 57.94  & 62.29  & 73.03 \\
0.5:0.01:1    & 59.25  & 62.14  & 73.61 \\
0.1:0.01:1    & 61.24  & 61.94  & 73.54 \\
\bottomrule
\end{tabular}
\caption{Performance under different $\gamma$ configurations on the CUB200 dataset.}\label{tab:gamma}
\end{table}

Hence, we can safely conclude that our OFCL framework exhibits admirable robustness across various configurations. The performance remains stable under various hyperparameter settings, suggesting that the model is not overly sensitive to the precise tuning of loss weights. This indicates that OFCL can maintain its effectiveness in real-world scenarios where hyperparameter optimization may be limited or costly.

Notably, while some degree of tuning is expected, our framework involves only 4–5 key hyperparameters, which constitutes a manageable level of complexity in practical deployments. Furthermore, we have open-sourced our implementation, and state-of-the-art performance can be achieved even with commonly used default settings. These results highlight both the robustness and ease of adoption of the proposed OFCL framework.

\section{Conclusions and Future Work}
In this paper, we proposed OFCL, a novel framework for open-world continual learning under limited labeled data. Technically, the core strength of OFCL lies in its dynamic adaptation of the knowledge space, which enables effective mining, representation, and continual accumulation of knowledge for both known and unknown classes. As a result, OFCL not only enhances the semantic representation of known categories, but also improves the compactness and separability of decision boundaries, thereby facilitating more accurate open-set recognition.

Furthermore, OFCL is inherently backbone-agnostic and can be seamlessly integrated with various embedding networks, making it a plug-and-play solution with strong generalization capability. Looking ahead, an important direction for future work is the integration of large language models (LLMs) into the framework to further enrich semantic understanding. Additionally, we aim to improve the interpretability of the learned representations and develop effective test-time adaptation techniques to enhance the model’s capability in handling previously unseen classes.


\section*{Acknowledgments}
This work was supported by the National Natural Science Foundation of China (Nos. 62476228, U2468207, 62176221), the Sichuan Science and Technology Program (No. 2024ZYD0180).


\bibliographystyle{ACM-Reference-Format}
\balance
\bibliography{sample-base}

\newpage
\section*{Appendix}
\subsection*{A. Notations}
In \autoref{tab:notations}, we introduce the notations we use throughout this paper.
\begin{table}[ht]
 \begin{center}
  \setlength{\tabcolsep}{1mm}{
  \begin{tabular}{c c}
\hline
\textbf{Notation} & \textbf{Explanation} \\ \hline
$t$ & The session $t$ \\

$CLS_{tr}^t$ & The set of training classes from $t$\\

$D^t_{tr}, D^t_{te}$ & Training and test sets of $t$ \\

$N$ &  \# of classes in each training set\\

$K$ &  \# of training samples of each class\\

$C$ &  Class $C$ \\

$\mathbf{x}^+, \mathbf{x}^-$ & Positive and negative sets of class $C$\\

$\alpha, \beta$ & Scaling factors for $\mathbf{x}^+ \text{and }\mathbf{x}^-$\\

$q_\sigma(\cdot)$ & A quantile function \\

$\mathbf{p}^t$ & The token set of $t$\\

$\mathbf{p}*$ & The set of matched tokens\\

$L_P$ &The length of each token\\

$D_e$ &The embedding dimension\\

$\mathbf{h'}^t = \mathbf{h}^t \oplus  \mathbf{p}^*$ & The augmented embedding\\

$f_{pr}$ & The pre-trained backbone\\

$f_{\theta}$ & The trainable classifier\\

$r_{C}$ & The learnable radius of class $C$ \\

$\mathbf{c}_{C}$ & The learnable centroid of class $C$ \\

$m$ & The margin\\

$(CLS^t_{tr},\mathbf{C}^{t},\mathbf{R}^{t})$ & Hyperspheres learned from $D^t_{tr}$\\

$(CLS_{open}^t,\overline{\mathbf{C}^{t}},\overline{\mathbf{R}^{t}})$ & Hyperspheres learned from unknowns\\

$\lambda$ & The trade-off parameter\\
\hline
\end{tabular}}
\caption{Notations and explanations.} 
\label{tab:notations}
 \end{center}
\end{table}

\subsection*{B. Baselines Descriptions}
To thoroughly validate our proposed OFCL, we compare it against a wide range of benchmarks on two widely used datasets. OFCL framework outperforms previous works, setting a new state-of-the-art performance.
We provide detailed descriptions of all the baselines, including \emph{Unknowns Detection Baselines} and \emph{Knowns Classification Baselines}. As mentioned before, unknown detection baselines include 8 representative OOD and OSR methods and known classification baselines include 15 challenging few-shot continual learning methods.

\paragraph{Unknowns Detection Baselines}
\begin{itemize}
    \item \emph{MSP} uses the maximum softmax probability of classifier predictions as the OOD score.
    \item \emph{KL} proposes a scalable framework for OOD detection, leveraging large-scale datasets and improved evaluation protocols to enhance robustness and generalization.
    \item \emph{SSD} is a unified self-supervised method that integrates diverse pretext tasks to improve representations.  
    \item \emph{MaxLogits} enhances OOD detection performance by utilizing the maximum logit score.
    \item \emph{Energy} assigns lower energies to ID data and higher energies to OOD data to create an energy gap for OOD detection.
    \item \emph{ViM} is an OOD detection method with matching virtual logits, improving the separation between in-distribution and out-of-distribution samples through a robust scoring mechanism. 
    \item \emph{KNN-based OOD} presents a deep nearest neighbor-based method for OOD detection, utilizing the local structure of feature spaces to identify anomalies. 
    \item \emph{NNGuide} is a nearest neighbor guidance strategy for OOD detection, enhancing the discriminative power of feature representations by leveraging proximity-based metrics.
\end{itemize}

\paragraph{Knowns Classification Baselines}
\begin{itemize}
    \item \emph{iCaRL} is an incremental classifier and representation learning method designed for FSCIL.
    \item \emph{TOPIC} is a model tailored for FSCIL, emphasizing effective adaptation to new classes with limited samples. 
    \item \emph{CEC} accommodates new classes while preserving previous knowledge for incremental learning.
    \item \emph{Meta-FSCIL} is a meta-learning-based approach, leveraging meta-learning principles to facilitate fast adaptation to new sessions.
    \item \emph{ALICE} employs an attention-based approach to efficiently learn new classes with limited training samples.
    \item \emph{FeSSSS} addresses the challenge of FSCIL by utilizing feature space segmentation for improved adaptation.
    \item \emph{Pro-KT} The Pro-KT integrates task-generic and task-specific prompt libraries to encode and transfer knowledge, while leveraging task-aware open-set boundaries to effectively identify unknowns.
    \item \emph{LIMIT} focuses on efficient knowledge transfer and adaptation to new classes while mitigating the impact on the existing knowledge.
    \item \emph{NC-FSCIL} is a neural clustering-based model, utilizing clustering techniques to enhance adaptation. 
    \item \emph{MCNet} leverages memory-augmented neural networks for effective knowledge retention.
    \item \emph{SoftNet} utilizes the soft parameter sharing to facilitate efficient learning of new classes.
    \item \emph{CoSR} is an FSCIL model that employs a collaborative self-regularization mechanism for effective adaptation while preserving previous knowledge.
\end{itemize}

\subsection*{C. Training Details}
We followed the general setting in the few-shot continual learning community, i.e., randomly shuffled the session order for both CUB200 (with 10 sessions) and MiniImageNet (with 8 sessions).
For our setting, we remove the labels of the classes in the training set of the next session (i.e., session ${(t+1)}$) and treat them as unknowns in the test set for the current session (i.e., session $t$). 
The results presented throughout this paper are the mean results of three random shuffles.
We implement all experiments on one NVIDIA GeForce-RTX-3090 GPU and the Pytorch library. 
Input images are resized to 224 x 224 and normalized to the range of [0,1]. The hyperparameter settings for each baseline are set according to the optimal combination reported in their papers, respectively.
All models are trained by the Adam optimizer with a batch size of 25 and a learning rate of 0.03. 
Our contribution also includes reproducible PyTorch implementations of our method and a large number of baselines under the open-world continual learning setting.

\begin{table*}[ht]
\centering
\begin{tabular}{l@{\hspace{8pt}}c@{\hspace{8pt}}c@{\hspace{8pt}}c@{\hspace{8pt}}c@{\hspace{8pt}}c@{\hspace{8pt}}c@{\hspace{8pt}}c@{\hspace{8pt}}c@{\hspace{8pt}}c@{\hspace{8pt}}c@{\hspace{8pt}}c@{\hspace{8pt}}c@{\hspace{8pt}}c@{\hspace{8pt}}}
\hline
\multicolumn{1}{c}{\multirow{2}{*}{Methods}}  & \multicolumn{11}{c}{$ACC_N$ in each task (\%) $\uparrow$}  & \multirow{2}{*}{PD$\downarrow$} \\ \cline{2-12}
\multicolumn{1}{c}{}        & 0 & 1 & 2 & 3 & 4 & 5 & 6 & 7 & 8 & 9 & 10 &  &  \\ \hline
                                OFCL          & 76.67 & 81.09 & 76.80 & 76.20 & 74.80 & 72.44 & 71.52 & 69.17 & 70.33 & 68.17 & 68.92 & 7.74 \\
                OFCL*          & 72.00 & 72.46 & 69.12 & 73.57 & 69.02 & 66.91 & 66.45 & 64.19 & 65.75 & 63.00 & 63.80 & 8.20 \\
                 \hline
Performance Drop & 4.67 & 8.63 & 7.68 & 2.63 & 5.78 & 5.53 & 5.07 & 4.98 & 4.58 & 5.17 & 5.12 & -0.46 \\ 
\hline
\end{tabular}
\caption{Averaged classification accuracy (\%) on CUB200 (10-way 5-shot).}
\label{tab:known-cub2}
\end{table*}

\begin{table*}[ht]
\centering
\begin{tabular}
{l@{\hspace{8pt}}c@{\hspace{8pt}}c@{\hspace{8pt}}c@{\hspace{8pt}}c@{\hspace{8pt}}c@{\hspace{8pt}}c@{\hspace{8pt}}c@{\hspace{8pt}}c@{\hspace{8pt}}c@{\hspace{8pt}}c@{\hspace{8pt}}}
\hline
\multicolumn{1}{c}{\multirow{2}{*}{Methods}}  & \multicolumn{9}{c}{$ACC_N$ in each task (\%) $\uparrow$}  & \multirow{2}{*}{PD $\downarrow$} \\ \cline{2-10}
\multicolumn{1}{c}{}        & 0 & 1 & 2 & 3 & 4 & 5 & 6 & 7 & 8 &  \\ \hline           
                 OFCL             & {96.00} & {91.65} & {89.17} & 84.90 & 86.46 & {87.18} & {87.57} & {85.71} & {87.12} & 10.88\\
               OFCL*             & {87.60} & 56.00 & 65.43 & 62.93 & 62.50 & 61.63 & {65.03} & {60.81} & {65.24} & {22.35}\\
                \hline
Performance Drop & 8.40 & 35.65 & 23.74 & 21.97 & 23.96 & 25.55 & 22.54 & 24.90 & 21.88 & -11.47\\
\hline
\end{tabular}
\caption{Averaged classification accuracy (\%) on MiniImageNet (5-way 5-shot).}
\label{tab:known-mini2}
\end{table*}
\subsection*{D. Additional Experimental Results}

Due to the inability of existing incremental learning baselines to replicate under the OFCIL setting, we further demonstrate the capability of our proposed framework in open-world continual learning scenarios. Specifically, we implement the OFCL framework within this open-world continual learning setting and report the outcomes, referred to as OFCL*.
Unlike the results presented in the main text, the experimental results for OFCL* account for the impact of open detection. To provide a clearer comparison, we present the results of OFCL* alongside those of all baselines in \autoref{tab:known-cub2} and \autoref{tab:known-mini2}.

\autoref{tab:known-cub2} and \autoref{tab:known-mini2} showcase the OFCL* experimental results on the CUB200 and MINI datasets, respectively.
The results presented for comparative models are not influenced by unknown samples, whereas the OFCL* under the OFCIL setting is affected by unknown samples. As evident from \autoref{tab:known-cub2}, even the impacted OFCL* manages to achieve optimal performance in the majority of sessions. In \autoref{tab:known-mini2}, it is apparent that the OFCL* initially experiences a significant impact from unknown samples. However, with an increasing number of learning sessions, the OFCL* gradually surpasses other comparative models. From these observations, we can conclude that the OFCL* demonstrates strong adaptability to long sequential tasks, aligning with the conclusions drawn in the main text.

\subsection*{E. Visualization and Analysis}

To show the effectiveness of OFCL in learning the open decision boundary, we compare the visualizations of the final embeddings of all test samples in an arbitrary session using the vanilla backbone and our proposed OFCL framework. 

\begin{figure}[h]
    \centering
    \includegraphics[width=\linewidth]{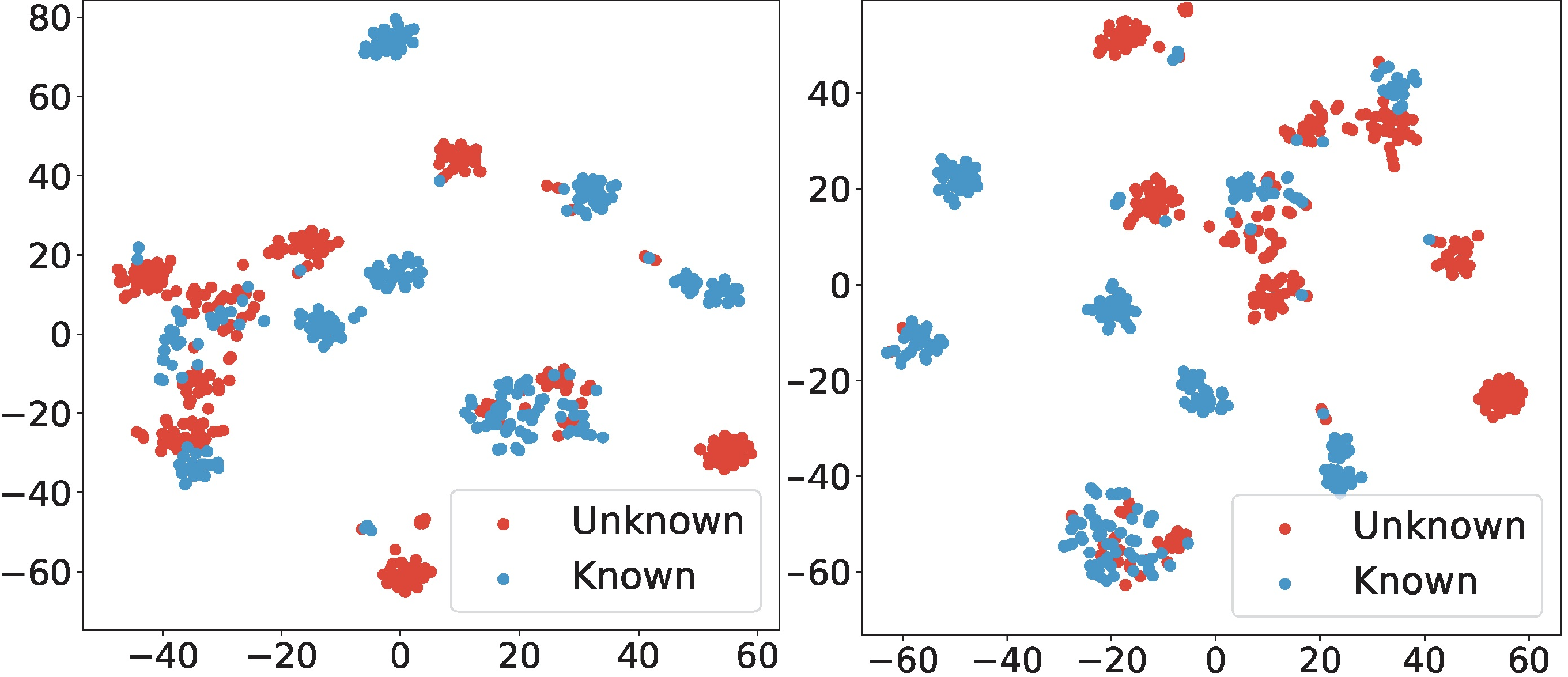}
    \caption{Visualization of the representations of all test samples using the vanilla Transformer-based backbone (left) and our proposed OFCL framework (right).}
    \label{fig:visual}
\end{figure}

As illustrated from the left part of \autoref{fig:visual}, we observe that there is no clear boundary between known points and unknown points, making it difficult to distinguish between them effectively.
In the right part of \autoref{fig:visual}, we can observe distinct boundaries between the known points and unknown points, along with a more balanced distribution of clusters. These visualizations provide compelling evidence that our proposed OFCL significantly improves the learning of discriminative and compact decision boundaries for open detection.

\end{document}